\documentclass[letterpaper, 10 pt, conference]{ieeeconf}  % Comment this line out if you need a4paper

\IEEEoverridecommandlockouts                              % This command is only needed if 
\usepackage{graphics} % for pdf, bitmapped graphics files
\usepackage{epsfig} % for postscript graphics files
\usepackage{mathptmx} % assumes new font selection scheme installed
\usepackage{times} % assumes new font selection scheme installed
\usepackage{amsmath} % assumes amsmath package installed
\usepackage{amssymb}  % needed for \mathbb in Problem Formulation
\usepackage{graphicx} % for \resizebox in the results table
\usepackage{booktabs} % for \toprule/\midrule/\bottomrule in tables
\usepackage{multirow} % for multi-row table cells
\usepackage{makecell} % for \makecell in table headers
\usepackage[table]{xcolor} % for \rowcolor shading in tables
\usepackage{wrapfig} % for the force-analysis wrapfigure

\let\IEEEoriginalthanks\thanks
\renewcommand{\thanks}[1]{%
  \IEEEoriginalthanks{\fontsize{6.5}{7}\selectfont #1}%
}

\title{\LARGE \bf
Collision-Aware Humanoid Whole-Body Control \\under Imperfect Tracking Targets
}

\author{Mohitvishnu S. Gadde$^{1}$, Ashish Mailk$^{1}$, Pranay Dugar$^{1}$, Aayam Kumar Shrestha$^{1}$, Fuxin Li$^{1}$ and Alan Fern$^{1}$
\thanks{This work is supported by NSF Award 2321851, DARPA contract HR0011-24-9-0423, and the NVIDIA Academic Grant Program.}
\thanks{$^{1}$All authors are associated with the Collaborative Robotics and Intelligent Systems Institute (CoRIS), Oregon State University, Corvallis, OR 97331. {\ttfamily \{gaddem, malikas, dugarp, aayam.shrestha, fuxin.li, afern\}@oregonstate.edu}}}

\begin{document}

\maketitle
\thispagestyle{empty}
\pagestyle{empty}

%%%%%%%%%%%%%%%%%%%%%%%%%%%%%%%%%%%%%%%%%%%%%%%%%%%%%%%%%%%%%%%%%%%%%%%%%%%%%%%%
\begin{abstract}

Humanoid robots often execute motion commands through whole-body controllers (WBCs) that track targets while maintaining balance and stability. However, most WBCs are blind to scene geometry, which can lead to collisions from imperfect target motions that are geometrically unsafe due to perception, planning, or teleoperation errors. We propose RECAL, a Robot--Environment Cross-Attention Layer that wraps a blind WBC to trade off target tracking against collision avoidance using external scene geometry. RECAL supports collision-aware tracking of floating-base and end-effector commands, including collision avoidance for held objects. It represents the robot, held objects, and environment as point clouds, using cross-attention between robot/object points and the environment to produce geometry-aware control features. In simulation, RECAL improves collision avoidance while preserving target-tracking performance across frozen-arm and adaptive-arm locomotion, object-carrying, and standing-manipulation scenarios relative to alternative geometry-aware WBC architectures. We further demonstrate the controller on a real Digit V3 humanoid robot.

\end{abstract}

%%%%%%%%%%%%%%%%%%%%%%%%%%%%%%%%%%%%%%%%%%%%%%%%%%%%%%%%%%%%%%%%%%%%%%%%%%%%%%%%
\section{INTRODUCTION}

A core component of many humanoid robot control architectures is the whole-body controller (WBC), which translates motion targets from teleoperation systems, planners, or learned skills into motor commands while maintaining balance and robustness. In cluttered environments, however, these targets may be imperfect as they can specify the intended behavior without fully defining a collision-free whole-body motion. This may occur due to partial targets, perception errors, planning approximations, or teleoperation imprecision. Since many WBCs are blind to the environment's geometry, naively tracking such targets can lead to unintended collisions involving the robot body, manipulated objects, and nearby obstacles.

We address this problem by learning a collision-aware control layer that enables an otherwise blind WBC to leverage local scene geometry when interpreting imperfectly tracked targets. Rather than requiring upstream systems to provide complete, collision-free whole-body references, the controller treats these targets as guidance and resolves local geometric conflicts at the control level. This allows high-level interfaces, such as learned skills, planners, and teleoperation, to operate more robustly in cluttered environments. For example, a teleoperator need not command every body part through a narrow doorway or specify a fully collision-free body motion when reaching into a constrained shelf.

Prior work has explored collision-aware humanoid control using control-theoretic and optimization methods (e.g., \cite{khatib2008unified,sentis2006whole,kumagai2018efficient}), but scaling such approaches to cluttered, high-dimensional humanoid settings remains challenging due to modeling and computational complexities. Recent learning-based controllers for humanoids incorporate terrain awareness (e.g., \cite{duan2024learning,sun2025learning,gadde2025no}), with related point-to-point controllers accounting for terrain and 3D obstacles \cite{xue2026collision,ben2025gallant}. Other works address collision avoidance for stationary humanoid end-effector control using time-of-flight sensors on the robot body \cite{kim2024armor,kohlbrenner2026egocentric}. To our knowledge, prior learning-based WBCs have not jointly addressed free-base locomotion, end-effector tracking, and collision avoidance for the robot body and held objects, which are required for robust loco-manipulation in cluttered environments.

To this end, we introduce the \emph{Robot--Environment Cross-Attention Layer (RECAL)}, a learned environment-aware wrapper around a blind WBC. RECAL modifies the behavior of the underlying controller to track base root and end-effector targets while avoiding collisions between the robot and the environment, and between objects and the environment. It represents the environment, robot body, and held-object geometry using point clouds derived from egocentric vision and proprioception. Robot and object points query the environment point cloud, enabling cross-attention to compute local, geometry-aware features for the robot and the manipulated object. RECAL is trained using a teacher--student distillation framework, where an observation-only student is distilled from the privileged information provided by the teacher policy.

We evaluate RECAL in simulation across cluttered environments and varying degrees of tracking imperfections. The experiments show that RECAL preserves accurate target tracking while avoiding collisions that occur when only using the blind WBC. We further show that the robot--environment cross-attention architecture outperforms alternative scene-geometry encoders. Finally, we demonstrate sim-to-real transfer on a Digit V3 humanoid robot.

Our main contributions are: {\bf 1) A collision-aware humanoid WBC framework} that tracks floating-base and end-effector targets while avoiding collisions with the environment. {\bf 2) The Robot Environment Cross-Attention Layer (RECAL)} that uses robot and held-object points as queries into the environment point cloud to produce collision-aware commands for an underlying blind WBC. {\bf 3) Simulation and real-robot validation} showing that RECAL can deviate from unsafe commands, avoid collisions, and recover target-tracking behavior across loco-manipulation scenarios.

%%%%%%%%%%%%%%%%%%%%%%%%%%%%%%%%%%%%%%%%%%%%%%%%%%%%%%%%%%%%%%%%%%%%%%%%%%%%%%%%

\begin{figure*}[t]
    \centering
    \includegraphics[width=0.9\linewidth]{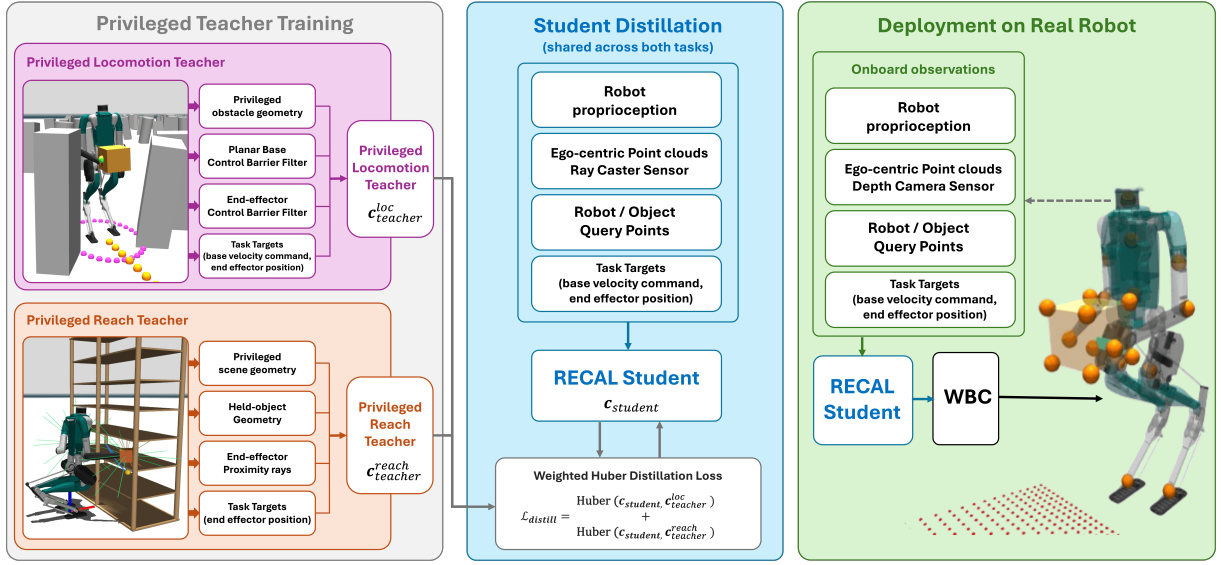}
    \caption{System overview: RECAL augments a blind WBC with geometry-aware command adaptation. Privileged teachers generate safe commands for locomotion and reaching, which are distilled into a shared set of student commands. At deployment, RECAL uses only onboard proprioception, egocentric point clouds, and robot/object query points to modify target commands before the WBC, avoiding unsafe motion while preserving feasible tracking.}
    \label{figure:architecture}
    \label{fig:RECAL}
    \vspace{-1em}
\end{figure*}

\section{RELATED WORK}
\label{sec:related_work}

\textbf{Classical and optimization-based collision-aware control.}
Classical collision-aware control builds on artificial potential fields, operational-space control, and constrained whole-body optimization. Potential-field methods give reactive avoidance for manipulators~\cite{khatib1986real}, while operational-space and task-priority formulations extend this to humanoids with many simultaneous objectives, contacts, and constraints~\cite{sentis2006whole,khatib2008unified}. Other works fold whole-body collision constraints into locomotion planning~\cite{kumagai2018efficient}. These methods are structured and interpretable, but typically rely on accurate geometric models, hand-designed constraint sets, and online optimization over high-dimensional humanoid states, which is difficult to scale to cluttered, contact-rich scenes.

\textbf{Learning-based and perceptive humanoid control.}
Learned humanoid WBCs can track motion, pose, or end-effector commands for whole-body loco-manipulation~\cite{he2024hover,li2024exbody2,dugar2025learning,he2025asap,dugar2025no,wang2025falcon,jiang2025gmt}, but are typically blind and rely on upstream targets to be collision-free. Perceptive humanoid controllers incorporate egocentric vision, RGB-D, LiDAR, proximity sensing, or visual imitation for navigation, terrain traversal, teleoperation, reaching, and local avoidance~\cite{merel2018hierarchical,merel2019catch,hansen2024hierarchical,yang2024mobile,ze2025twist2,luo2025perceptive,allshire2025videomimic,yin2025visualmimic,he2025viral,duan2024learning,sun2025learning,li2025clone,kim2024armor,kohlbrenner2026egocentric}. Closest to our setting are collision-aware locomotion policies for 3D-constrained spaces that use voxelized scene grids~\cite{ben2025gallant} or hand-designed potential-field features queried at body parts~\cite{xue2026collision}. However, existing approaches typically focus either on locomotion with limited upper-body objectives or on stationary reaching, and do not jointly address floating-base motion, end-effector tracking, and held-object collision avoidance. RECAL targets this combined setting by adapting floating-base, end-effector, arm, and held-object motions to avoid unintended contacts while preserving the command-tracking interface of the underlying WBC.

\textbf{3D geometry representations for control.}
A collision-aware layer must encode scene geometry so the controller can decide which controllable surface is near which obstacle. Point-based networks process unordered sets~\cite{qi2017pointnet}, voxel encoders impose structured grids~\cite{zhou2018voxelnet}, and point-transformer models give expressive encodings~\cite{wu2024pointtransformerv3}, but these global scene encodings do not directly reveal robot-obstacle relations. Robot-aware methods address this by binding controlled geometry to the scene: learned collision models query object poses against point clouds~\cite{danielczuk2021object}, neural motion planners condition on observed geometry~\cite{fishman2022motion,dalal2024neuralmp}, and recent policies cross-attend robot state, demonstrations, or multiview observations with scene geometry~\cite{huang2025prism,jain2024vid2robot,goyal2023rvt}. Following this query-conditioned view, RECAL uses robot and held-object points as queries into the environment point cloud to produce local geometry-aware features. Our experiments compare against PointNet and Voxel encoders for scene representation.

\begin{figure*}[!t]
    \centering
    \includegraphics[width=0.81\linewidth]{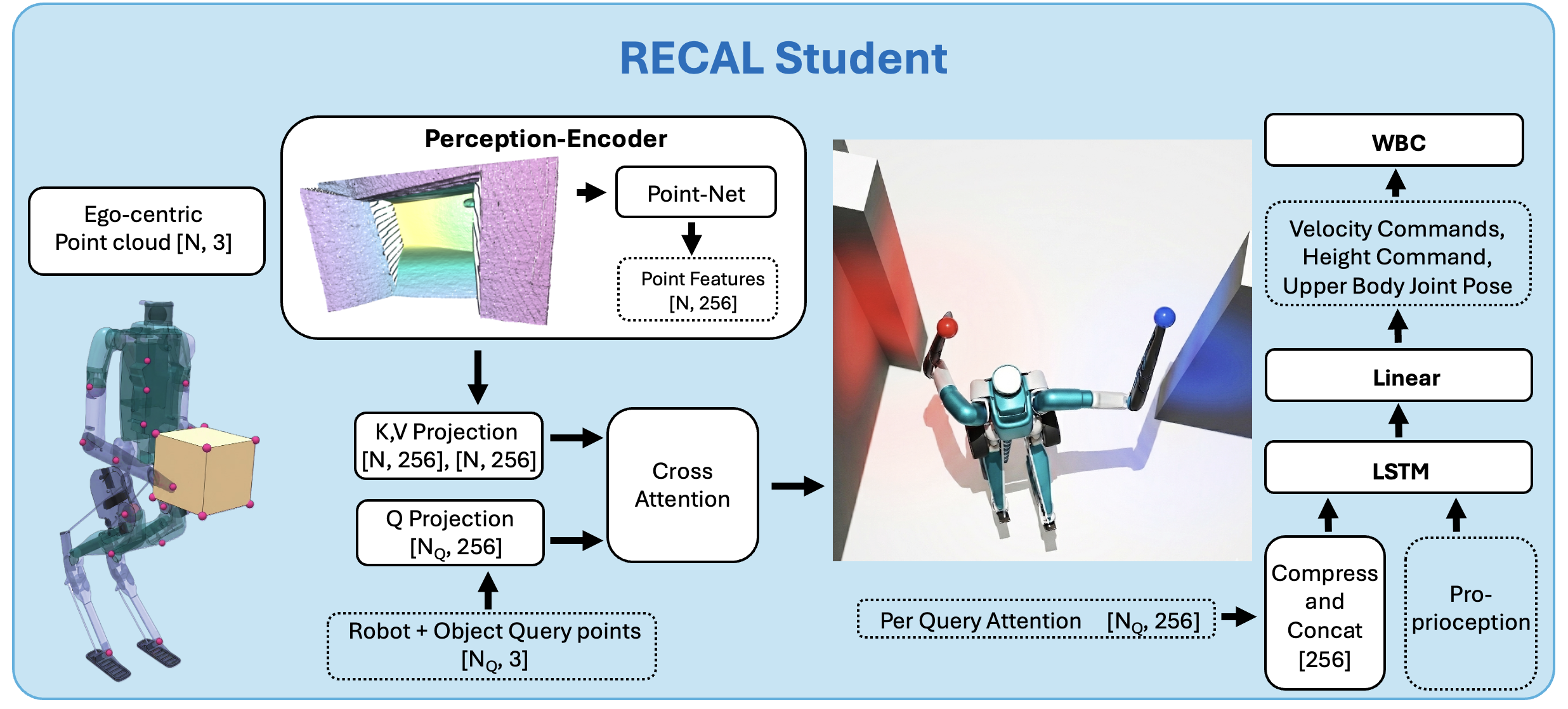}
    \caption{RECAL architecture. Environment points and robot/object query points are embedded and cross-attended to produce per-query, geometry-aware features. These features are compressed, concatenated with proprioception and the input command, and passed through a recurrent command adapter to produce the modified WBC command $\hat{c}_t$.}
    \label{fig:recal_arch}
    \vspace{-1em}
\end{figure*}

%%%%%%%%%%%%%%%%%%%%%%%%%%%%%%%%%%%%%%%%%%%%%%%%%%%%%%%%%%%%%%%%%%%%%%%%%%%%%%%%

\section{PROBLEM FORMULATION}
\label{sec:problem_formulation}

We consider a humanoid robot operating in cluttered environments, with target motions generated by an upstream module. Our goal is to learn a collision-aware WBC for loco-manipulation in which targets may specify a combination of floating-base motions, end-effector motions or postures. This interface supports stationary manipulation, locomotion, and loco-manipulation. However, the targets under-specify the desired whole-body behavior, leaving many degrees of freedom to the controller. The controller must control whole-body motion to track targets as closely as possible while avoiding unintended contact with the environment. Although we focus on a loco-manipulation WBC interface, the formulation applies to other interfaces with varying motion specifications.

The inputs for the WBC include \emph{robot proprioceptive state} $s_t = (\theta_t, \dot{\theta}_t, \alpha_t, q_t)$, where $\theta_t \in \mathbb{R}^J$ and $\dot{\theta}_t \in \mathbb{R}^J$ are the joint positions and velocities, $\alpha_t \in \mathbb{R}^3$ is the base angular velocity, and $q_t \in \mathbb{R}^4$ is the base orientation. The robot observes the environment through an egocentric point cloud $\mathcal{P}_t \subset \mathbb{R}^3$. Since $\mathcal{P}_t$ represents the geometry to avoid, points corresponding to intended contact surfaces can be removed before being input to the WBC. Finally, to represent the objects supported by the robot, we let $b_t^L$ and $b_t^R$ denote the bounding boxes associated with the left and right end-effectors, respectively. When an end-effector is not supporting an object, the corresponding box is set to a default end-effector collision box, encoding both held-object and end-effector geometry.

A target command for the WBC interface at time $t$ is denoted by $c_t = (v_t, \omega_t, \eta_t, e_t^R, e_t^L, f_t)$, where $v_t \in \mathbb{R}^2$ is the desired planar base velocity, $\omega_t \in \mathbb{R}$ is the desired turn rate, $\eta_t$ denotes desired base orientation parameters, and $e_t^R, e_t^L \in \mathbb{R}^3$ are desired right and left end-effector positions expressed in the base frame. The binary flag $f_t \in \{0,1\}$ specifies whether the end-effector poses should be frozen during locomotion. With $f_t=1$, the controller strongly prioritizes preserving the commanded end-effector positions and avoids collisions primarily through adjustments to the base motion. Otherwise, the controller may freely adjust the end-effector positions to avoid collisions. This distinction supports loco-manipulation behaviors in which the robot carries an object and must maintain a fixed grasp or object pose, as well as behaviors in which the arms may move freely.

At each time step $t$, the collision-aware WBC receives the current proprioceptive state $s_t$, environment point cloud $\mathcal{P}_t$, left and right end-effector bounding boxes $b_t^L, b_t^R$, and target command $c_t$. It outputs an action $a_t = \pi^{\mathrm{CA}}(s_t, \mathcal{P}_t, b_t^L, b_t^R, c_t)$ corresponding to PD setpoints for the $J$ actuated joints. The objective is to track the commanded targets as closely as possible while avoiding collisions with the environment, the robot body, end-effectors, and/or held-object bounding boxes.

%%%%%%%%%%%%%%%%%%%%%%%%%%%%%%%%%%%%%%%%%%%%%%%%%%%%%%%%%%%%%%%%%%%%%%%%%%%%%%%%

\section{COLLISION-AWARE WBC ARCHITECTURE}
\label{sec:architecture}

Our approach builds a collision-aware humanoid controller by layering a Robot--Environment Cross-Attention Layer (RECAL) on top of an existing, pre-trained blind WBC, as illustrated in Figure~\ref{fig:RECAL}. It provides robust balance, locomotion, and command tracking, but is blind to scene geometry. RECAL adds collision-aware command adaptation by using sparse robot and held-object query points to attend to the environment point cloud, producing a modified command $\hat{c}_t$ that trades off target tracking against local collision avoidance. The base WBC then maps $\hat{c}_t$ to joint-level PD targets. This layered design avoids training a new WBC from scratch, and instead leverages the stability, tracking behavior, and sim-to-real robustness of the underlying blind controller while learning to adjust its inputs to respond to local scene geometry.

\subsection{Base Whole-Body Controller.} We build on the pre-trained Masked Humanoid Controller (MHC)~\cite{dugar2025learning} as our base WBC, which maps kinematic target commands to joint-level PD position targets. The MHC has demonstrated robust sim-to-real transfer on the Digit V3 humanoid. It supports multiple control modes, and in this work, we use its loco-manipulation mode, which accepts floating-base motion commands together with arm joint targets. We denote the WBC as a fixed policy $a_t = \pi^{\mathrm{WBC}}(s_t,\hat{c}_t)$, where $s_t$ is the proprioceptive state and
$\hat{c}_t = [\hat{v}_t, \hat{\omega}_t, \hat{\eta}_t, \hat{u}_t]$
is the WBC target command. Here, $\hat{v}_t \in \mathbb{R}^2$ is the target planar base velocity, $\hat{\omega}_t \in \mathbb{R}$ is the target turn rate, $\hat{\eta}_t$ denotes target base orientation parameters, and $\hat{u}_t \in \mathbb{R}^8$ denotes arm joint targets, with four actuators per arm for Digit V3. The output $a_t$ specifies joint position PD targets for all $J=20$ actuated joints.

\subsection{RECAL Architecture.} RECAL is a policy, denoted $\pi^{\mathrm{REC}}$, that maps the collision-aware WBC inputs to modified commands for the blind base WBC:
\[
\hat{c}_t = \pi^{\mathrm{REC}}(s_t, \mathcal{P}_t, b_t^L, b_t^R, c_t).
\]
Composing RECAL with the blind WBC yields the collision-aware WBC:
\[
\pi^{\mathrm{CA}}(s_t, \mathcal{P}_t, b_t^L, b_t^R, c_t)
=
\pi^{\mathrm{WBC}}\!\Big(
s_t,
\pi^{\mathrm{REC}}(s_t, \mathcal{P}_t, b_t^L, b_t^R, c_t)
\Big),
\]
which returns PD setpoints for all robot joints. Intuitively, $\pi^{\mathrm{REC}}$ must be trained to adapt imperfect input commands into WBC-compatible commands that preserve the intended motion tracking when safe, but deviate from precise motion tracking when necessary to avoid collisions.

At a high level, RECAL consists of a robot-centric geometry encoder followed by a recurrent command adapter. The encoder maps the environment point cloud into features associated with sparse robot and held-object points, producing scene representations that are relevant to each robot and held-object points. These representations are compressed and concatenated with proprioceptive state and command information for an LSTM policy, which outputs the modified command $\hat{c}_t$.

The geometry encoder $E_\phi$ takes as input the egocentric environment point cloud $\mathcal{P}_t \in \mathbb{R}^{N \times 3}$ and a query point set $\mathcal{Q}_t \in \mathbb{R}^{M \times 3}$, all expressed in the robot base frame. The query set has a fixed layout with $M=37$ points divided into three groups: 13 anatomical body keypoints obtained using forward kinematics, 8 corners of a cuboid proxy for a bimanually carried object, and 16 hand-object queries corresponding to 8 cuboid corners for each end-effector. The body keypoints include three torso points, three points on each arm, and four leg points corresponding to the knees and feet. The object queries are not tied to a category and rather encode the occupied volume of arbitrary held objects using conservative cuboid proxies. Inactive object queries are masked. Together, these queries summarize the robot and held-object geometry whose collisions should be avoided.

The encoder first maps each environment point to point features using a shared MLP, and queries points using a separate MLP combined with learned query-identity embeddings to distinguish body regions and object-query types. The embedded robot and object queries then attend to environmental features via multi-head cross-attention. This produces one local context feature per query, encoding nearby scene geometry relative to the corresponding robot or object point: $Z_t = E_\phi(\mathcal{P}_t, \mathcal{Q}_t)$. The per-query outputs $Z_t$ are compressed by an MLP, and concatenated with the proprioceptive state $s_t$ and input command $c_t$. The resulting feature vector is projected to 256 dimensions and passed through a two-layer LSTM with 256 hidden units. A linear command head maps the LSTM output to the modified WBC command $\hat{c}_t$. The recurrent state smooths command corrections over time and helps compensate for partial or intermittent point cloud observations.

\section{RECAL TRAINING}
\label{sec:recal_training}

We train RECAL using teacher--student distillation. Since desirable command modifications depend on collision geometry, which is easier to access in simulation, we construct privileged teachers that observe task-specific geometric information and produce target-modified commands $\hat{c}^{\mathrm{tea}}_t$. The RECAL student is then trained to imitate these teacher commands using only observational inputs. We use two teachers corresponding to the two regimes: locomotion through clutter, optionally while carrying objects with frozen end-effectors, and stationary reaching, where the base remains fixed while one or both hands follow potentially imperfect end-effector targets near shelf-like structures.

\subsection{Training Environments.} The locomotion-scenarios training environment is adapted from~\cite{ben2025gallant}. It consists of a large area populated with obstacles of varying size and placement, with randomized robot locations, object-carrying state, end-effector freeze mode, and target command sequences $c_t$. For stationary-reaching scenarios, the robot is initialized in front of procedurally generated cabinets and open shelves, with randomized furniture dimensions, shelf counts, partitions, and per-hand object extents. Here, $c_t$ specifies end-effector trajectories to reach sampled locations in or around the furniture while keeping the base stationary. In both scenarios, $c_t$ may pass through obstacles, causing collisions if tracked blindly.

\subsection{Privileged Teachers.} We begin by training privileged teachers using simulator-only geometric information to produce corrected WBC commands. The teachers receive the same upstream command $c_t$ as the student and output a WBC-compatible target command $\hat{c}^{\mathrm{tea}}_t$. These teachers do not generate the task command itself, but supervise corrections to avoid unintended contact. We use two teachers with complementary forms of privileged supervision. For locomotion, object carrying, and frozen-arm navigation, we use an analytic teacher based on CBF-style safety filtering: it modifies the commanded base motion, and when allowed, the end-effector targets, using privileged obstacle geometry. When the robot carries an object, the teacher enlarges the safety region to account for the object's occupied volume. For stationary reaching, we use a privileged recurrent policy trained with RL to track end-effector targets while avoiding contact. This teacher observes simulator-only furniture geometry, hand-object extents, and upper-body contact signals. Both teachers' outputs corrected the commands in the shared WBC command representation.

For locomotion and object carrying, the analytic teacher applies CBF-style projections to the commanded planar base velocity using privileged obstacle geometry, enlarging the safety region when a held object is active. When end-effectors are not frozen, commanded hand targets are similarly displaced away from nearby obstacles and converted to arm-command components with damped least-squares IK. For stationary reaching, the privileged teacher is a recurrent PPO policy that observes furniture geometry, hand-object extents, local proximity rays, active-arm state, and upper-body contact signals, and outputs a root-height residual and two 4-D arm residuals in the shared WBC command representation.

\newcommand{\taskhead}[1]{\multicolumn{2}{c|}{\makecell[c]{#1}}}
\begin{table*}[!t]
\centering
\caption{
Performance across locomotion and reach tasks under varying difficulty levels.
CF Success $\uparrow$ denotes collision-free success rate; Dev $\downarrow$
denotes command deviation. Shaded rows indicate deployable student policies.
\textbf{Bold} marks the best student policy per task.
Averaged over 5 seeds.}
\label{tab:combined_results}

\scriptsize
\setlength{\tabcolsep}{3pt}
\renewcommand{\arraystretch}{1.03}
% \resizebox{\textwidth}{!}  % previous: stretched table (and font) to fill the full page width, making the font look too big
\resizebox{0.85\textwidth}{!}
{
\begin{tabular}{c | l | cc | cc | cc | cc | cc |}
\toprule

\multirow{2}{*}{\centering Difficulty} &
\multicolumn{1}{c|}{\multirow{2}{*}{Method}} &

\taskhead{Adaptive-arm\\Locomotion} &
\taskhead{Frozen-arm\\Locomotion} &
\taskhead{Object\\Carrying} &
\taskhead{Stationary\\Reach} &
\taskhead{Reach w/\\Held Object} \\

&
& Succ$\uparrow$ & Dev$\downarrow$
& Succ$\uparrow$ & Dev$\downarrow$
& Succ$\uparrow$ & Dev$\downarrow$
& Succ$\uparrow$ & Dev$\downarrow$
& Succ$\uparrow$ & Dev$\downarrow$ \\
\midrule
\midrule

& Blind-WBC
& 0.62 & 0.121
& 0.62 & 0.122
& 0.81 & 0.120
& 0.43 & 0.074
& 0.47 & 0.071 \\

& Teacher
& 0.99 & 0.147
& 0.93 & 0.121
& 1.00 & 0.123
& 0.99 & 0.022
& 0.95 & 0.030 \\

\rowcolor{gray!12}

Easy
& \textbf{Student (RECAL)}
& 0.95 & 0.184
& \textbf{0.91} & \textbf{0.144}
& \textbf{0.96} & \textbf{0.131}
& \textbf{0.95} & 0.034
& \textbf{0.90} & \textbf{0.039} \\

\rowcolor{gray!12}
& Student (PointNet)
& \textbf{0.96} & \textbf{0.172}
& 0.89 & 0.145
& 0.89 & 0.140
& 0.58 & 0.063
& 0.60 & 0.068 \\

\rowcolor{gray!12}
& Student (Voxel)
& 0.42 & 0.178
& 0.43 & 0.153
& 0.76 & 0.152
& 0.73 & \textbf{0.033}
& 0.34 & 0.050 \\

\midrule

& Blind-WBC
& 0.38 & 0.133
& 0.42 & 0.133
& 0.62 & 0.123
& 0.41 & 0.076
& 0.37 & 0.086 \\

& Teacher
& 0.97 & 0.181
& 0.96 & 0.134
& 1.00 & 0.135
& 0.93 & 0.025
& 0.89 & 0.037 \\

\rowcolor{gray!12}

Medium
& \textbf{Student (RECAL)}
& \textbf{0.93} & 0.213
& \textbf{0.93} & \textbf{0.159}
& \textbf{0.98} & \textbf{0.144}
& \textbf{0.86} & 0.036
& \textbf{0.78} & \textbf{0.054} \\

\rowcolor{gray!12}
& Student (PointNet)
& 0.83 & \textbf{0.201}
& 0.85 & 0.165
& 0.83 & 0.146
& 0.55 & 0.068
& 0.45 & 0.094 \\

\rowcolor{gray!12}
& Student (Voxel)
& 0.25 & 0.231
& 0.33 & 0.173
& 0.55 & 0.161
& 0.48 & \textbf{0.031}
& 0.39 & 0.061 \\

\midrule

& Blind-WBC
& 0.32 & 0.140
& 0.33 & 0.141
& 0.51 & 0.127
& 0.45 & 0.080
& 0.39 & 0.089 \\

& Teacher
& 0.96 & 0.205
& 0.89 & 0.153
& 1.00 & 0.159
& 0.87 & 0.035
& 0.81 & 0.051 \\

\rowcolor{gray!12}

Hard
& \textbf{Student (RECAL)}
& \textbf{0.91} & \textbf{0.193}
& \textbf{0.86} & 0.178
& \textbf{0.92} & 0.155
& \textbf{0.81} & 0.041
& \textbf{0.72} & \textbf{0.064} \\

\rowcolor{gray!12}
& Student (PointNet)
& 0.65 & 0.270
& 0.56 & 0.176
& 0.63 & \textbf{0.155}
& 0.47 & 0.079
& 0.43 & 0.094 \\

\rowcolor{gray!12}
& Student (Voxel)
& 0.21 & 0.221
& 0.31 & \textbf{0.170}
& 0.52 & 0.162
& 0.37 & \textbf{0.037}
& 0.29 & 0.073 \\

\bottomrule
\end{tabular}
}
\vspace{-1.5em}
\end{table*}

\subsection{Student Distillation.} We train a single RECAL student to imitate both privileged teachers using deployable observation inputs. The student predicts $\hat{c}_t = \pi^{\mathrm{REC}}(s_t,\mathcal{P}_t,b_t^L,b_t^R,c_t)$, which is supervised with teacher produced target commands, $\hat{c}^{\mathrm{tea}}_t$, as:
\[
    \mathcal{L}_{\mathrm{distill}}
    =
    \mathrm{Huber}_\beta\!\left(\hat{c}_t, \hat{c}^{\mathrm{tea}}_t\right),
    \qquad \beta=0.1.
\]
To emphasize rare but important corrective events, we upweight the loss on base-command components when the locomotion teacher deflects from the upstream base command, and we upweight the loss on arm-command components when either teacher moves an end-effector away from an obstacle. The recurrent student is trained with truncated backpropagation through time using chunks of length 64 and a cosine learning-rate schedule from $10^{-4}$ to $10^{-5}$. The fixed query layout with masking, as described in Section~\ref{sec:architecture}, indicates which robot or held-object geometry is active, allowing a single student to cover locomotion, object-carrying, and reaching tasks.

%%%%%%%%%%%%%%%%%%%%%%%%%%%%%%%%%%%%%%%%%%%%%%%%%%%%%%%%%%%%%%%%%%%%%%%%%%%%%%%%

\section{EXPERIMENTAL RESULTS}
\label{sec:result}

\subsection{Baselines.} Our experiments compare RECAL against the following baselines. \emph{1) Blind} is the underlying blind WBC directly executing the target command, intended as a lower performance bound. \emph{2) Teachers} are the privileged teachers (Section~\ref{sec:recal_training}) used to train the non-privileged students and are intended as an upper-bound reference. \emph{3) PointNet} replaces RECAL's query-conditioned cross-attention encoder with a PointNet~\cite{qi2017pointnet} encoder over environment, robot, and object points. \emph{4) Voxel} replaces the RECAL encoder with the voxel-based encoder from recent work on environment-aware humanoid locomotion~\cite{ben2025gallant}. The learned student baselines use the same recurrent policy, teacher labels, and distillation setup as RECAL; only the geometry encoder and its corresponding observation representation are changed.

\subsection{Evaluation Protocol and Metrics.} We evaluate RECAL on target commands that can be tracked well in free space but become unsafe in cluttered environments. We evaluate this behavior across locomotion-only, locomotion with frozen upper-body arm joints, moving while carrying objects, stationary reaching, and reaching with held objects. Furthermore, each task is evaluated across three levels (\emph{Easy, Medium, Hard}) of geometric difficulty, corresponding to increasingly imperfect target commands: locomotion tasks use decreasing reference-trajectory clearance margins, while reaching tasks use decreasing shelf-clearance gaps.

We report two main metrics. \emph{Collision-free success (CF)} is the fraction of episodes completed without unintended robot--environment or object--environment contact, computed over the full episode. \emph{Tracking deviation (Dev)} measures command-tracking error on safe trajectory states, so controllers are not penalized for tracking error during contact or near-contact states. For locomotion, \emph{Dev} is the planar velocity error between achieved and commanded body velocity; for reaching, \emph{Dev} is the Euclidean hand-to-target distance.

\begin{figure*}[ht]
    \centering
    \includegraphics[width=0.9\linewidth]{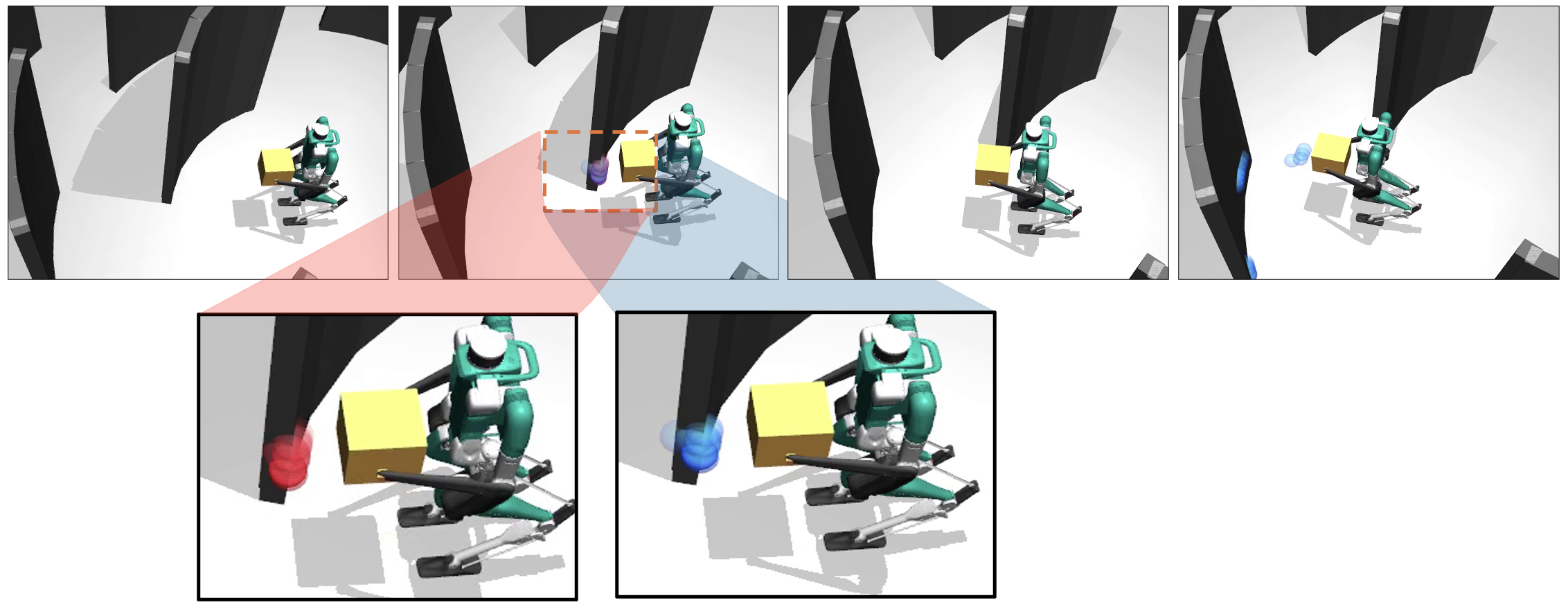}
    \caption{Evolution of hand query attention scores during a locomotion task execution. We show the attention scores for the local environment point cloud at left-hand (blue) and right-hand (red) query points. As the hands approach an obstacle, they attend more strongly to nearby relevant geometry, producing local geometry-aware features for collision avoidance.}
    \label{fig:attention}
    % \vspace{-1.5em}
\end{figure*}

\subsection{Locomotion and Object Carrying Performance.} Table~\ref{tab:combined_results} reports results for adaptive-arm locomotion, frozen-arm locomotion, and box carrying. As expected, the CF gap between Blind and Teacher grows with task difficulty, indicating that intentional collision avoidance is necessary. Blind often achieves lower Dev than Teacher because Dev is computed only over collision-free portions of trajectories, and when Blind does not collide, it remains an effective tracker. Across all difficulty levels, RECAL substantially improves CF over Blind and approaches Teacher performance. For example, at medium difficulty, RECAL reaches 0.93 CF versus 0.38 for Blind and 0.97 for Teacher, with only a modest Dev increase in harder settings.

RECAL also outperforms the alternative geometry encoders in CF while maintaining comparable Dev. PointNet is the strongest baseline and is competitive on easy tasks, but the gap widens as difficulty increases. For example, in hard adaptive-arm locomotion, PointNet reaches 0.65 CF, while RECAL reaches 0.91. This suggests that global scene features are less effective in tight settings, where collision avoidance depends on which robot or held object points are near which obstacles.

Voxel performs substantially worse in the CF metric across difficulty levels. One likely explanation is that this more general scene encoder must learn robot--environment relationships indirectly from global scene embeddings, robot points, and proprioception, whereas RECAL explicitly cross attends between the environment, robot, and held-object points. Although longer training or additional hyperparameter tuning may improve this baseline, it was trained for over $5\times$ as many steps as RECAL. Overall, these results support the value of RECAL's robot-centric cross-attention inductive bias.

\subsection{Stationary Reaching Performance.} Table~\ref{tab:combined_results} reports results for stationary reaching with and without held objects. Here, the collision avoidance depends entirely on upper-body reconfiguration: the arms, torso, and any held object must navigate constrained shelf geometry that the upstream end-effector target does not fully specify. As in locomotion, the large CF gap between Blind and Teacher shows the value of explicit collision avoidance. Hard reaching tasks are more challenging than locomotion, especially with held objects, where Teacher reaches 0.81 CF, due to tighter clearances and higher precision demands. Unlike locomotion, Blind shows higher Dev than Teacher because some reaching commands, especially large height changes, fall outside the WBC's effective command range; the RL Teacher compensates by adjusting the targets.

RECAL substantially improves performance over Blind and remains much closer to Teacher. It also outperforms the alternative encoder baselines, even at the easiest difficulty level. The gap is particularly pronounced for reaching compared to locomotion because the task requires fine-grained robot-scene reasoning. A global PointNet encoding can summarize shelf geometry but does not directly encode which hand, arm segment, or object corner is in a near-collision state. Voxel encodings preserve spatial structure, but fixed grid resolution makes it difficult to represent narrow shelf clearances and small object margins without high computational cost. In contrast, RECAL provides each hand, body, and held-object query with a local obstacle-aware feature, enabling collision reasoning across both robot parts and object volumes.

\textbf{Performance Summary.} These results show that RECAL improves collision-free execution across both locomotion-like and reaching-like interfaces. The gains are largest when the target command is most geometrically ambiguous: frozen arms, held objects, and tight reaching spaces. This supports RECAL's inductive bias: binding scene geometry to robot and object queries enables the policy to preserve feasible motion while modifying unsafe commands.

\subsection{Attention Visualization.} To illustrate the role of RECAL's cross-attention mechanism, Figure~\ref{fig:attention} visualizes attention weights for the left- and right-hand queries during a locomotion trajectory around obstacles. When the robot is far from relevant obstacles, attention is largely directed toward the global sink token rather than toward environment points. As the robot approaches an obstacle, attention shifts to the relevant obstacle points, enabling a local encoding of nearby scene geometry. In the fourth frame, attention is assigned to a wall that is still relatively distant but lies along the robot's direction of motion, suggesting that RECAL attends to geometry based on both proximity and motion context. We observe similar patterns for other robot queries and scenarios. For example, during object carrying, held-object corner queries attend to obstacles that may be far from the hands but would intersect the swept volume of the carried object.

\subsection{Force Analysis.} 
\begin{figure}
    % \vspace{-2em}
    \centering
    \includegraphics[width=0.98\linewidth]{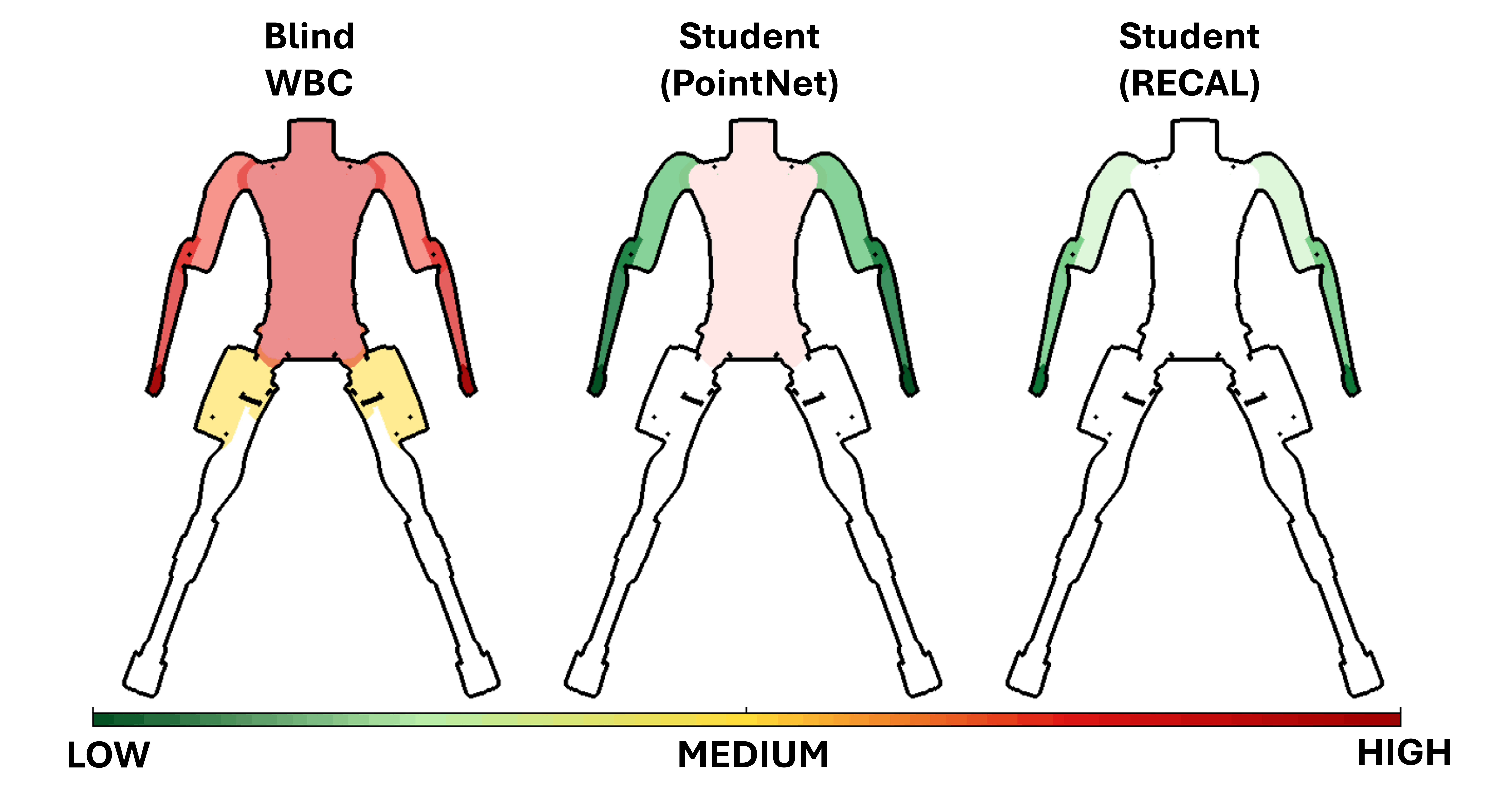}
    \caption{ Averaged contact force on the robot body for \emph{Hard} variant of Adaptive-arm locomotion, Frozen-arm locomotion and Stationary reach tasks. RECAL achieves significantly lower collision frequency and impact forces on the robot body.}
    \label{fig:forces}
    % \vspace{-2em}
\end{figure}

In practice, collisions vary in severity and meaning: a brief, arm brush is not equivalent to a sustained torso collision. To capture this, Figure~\ref{fig:forces} reports the average collision-force magnitude for different robot body regions across the hardest task settings. Blind experiences relatively severe torso collisions and moderate hip collisions. PointNet substantially reduces torso forces and further reduces arm forces. RECAL nearly eliminates torso collisions, with the remaining contacts primarily on the arms and at magnitudes far below Blind and comparable to PointNet. This reduction occurs while RECAL also substantially outperforms Blind and PointNet in collision-free success.

This analysis reinforces the need to interpret tracking deviation alongside collision-free success. Some baselines maintain low deviation by continuing to track target commands through contact, leading to larger collision forces. RECAL often allows more deviation near obstacles, but this deviation corresponds to physically meaningful avoidance that reduces collision severity while preserving tracking when avoidance is unnecessary.

% \textbf{Physical Robot Demonstration.} The supplementary material and video provide examples of RECAL controlling a Digit v3 humanoid robot. Command  targets are given to RECAL through a teleoperation interface for both locomotion and stationary-reaching tasks.

%%%%%%%%%%%%%%%%%%%%%%%%%%%%%%%%%%%%%%%%%%%%%%%%%%%%%%%%%%%%%%%%%%%%%%%%%%%%%%%%
\subsection{Physical Robot Demonstration}
\label{sec:physical_demo}

We demonstrate RECAL on a physical Digit V3 humanoid robot in both locomotion and stationary-reaching settings. For real-world perception, we mount two Intel RealSense D455 depth cameras on the robot, placed at approximately chest and pelvis height. The depth outputs from these cameras are converted into an egocentric environment point cloud expressed in the robot frame, which is ingested directly by the RECAL student. The robot query points are computed online by performing forward kinematics on the current robot state at each control step, whereas held-object query points are interpolated from known object dimensions. Together with proprioception and the target command, these query points enable RECAL to reason about the local relationships among the robot, any held object, and nearby scene geometry. Policy inference is performed on a computer equipped with an AMD Ryzen 9 9950X 16-core processor and an NVIDIA RTX 5090 GPU. The resulting PD commands are transmitted to the robot over UDP.

During demonstrations, locomotion commands are provided via a teleoperation interface, and the same end-effector command generator used during training is used for the stationary-reaching tasks. RECAL adapts these commands before passing them to the blind WBC. The robot can deviate from geometrically unsafe teleoperated commands near obstacles while maintaining stable whole-body behavior, then return to the intended tracking behavior once the motion becomes feasible. We successfully demonstrate all five tasks---adaptive arm locomotion, frozen-arm locomotion, object carrying, stationary reaching, and reaching while holding an object. In a few scenarios, the robot also makes light contact with the environment while executing the commanded motion. Please refer to the attached video.

These demonstrations provide qualitative evidence that RECAL transfers beyond simulation and can operate as a real-time safety-aware command-adaptation layer around an existing humanoid WBC.

\section{SUMMARY AND LIMITATIONS}
\label{sec:summary_limitations}

We introduced RECAL, a learned layer that wraps a frozen blind WBC and modifies its input commands to track intended motions while avoiding collisions with scene geometry. RECAL uses robot and held-object query points to cross-attend to an egocentric environment point cloud, producing local geometry-aware features for collision-aware command adaptation. Distilled from privileged locomotion and reaching teachers into a single observation-only student, RECAL substantially improves collision-free success over the blind base controller while preserving free-space tracking and approaching teacher performance across difficulty levels. The RECAL design outperforms PointNet and Voxel encoders, and transfers to a real Digit V3 humanoid.

Several limitations remain. RECAL currently assumes static, flat-ground environments and relies on egocentric depth observations with limited angular coverage, leaving some overhead and rear obstacles unobserved. It also uses a purely geometric avoidance formulation, treating perceived geometry as forbidden contact and therefore cannot reason about intentional or functional contact, semantic differences between surfaces, or movable obstacles. The learned behavior is limited by the privileged teachers used for distillation, which do not cover the full space of collision-aware strategies, such as more effective use of body rotation during locomotion. Finally, our evaluations are limited to rigid cuboid-held objects and non-dexterous end-effectors, and the policy does not yet emit an infeasibility signal when no collision-free execution exists.

% \section{CONCLUSIONS}

% A conclusion section is not required. Although a conclusion may review the main points of the paper, do not replicate the abstract as the conclusion. A conclusion might elaborate on the importance of the work or suggest applications and extensions. 

\addtolength{\textheight}{-12cm}   % This command serves to balance the column lengths

%%%%%%%%%%%%%%%%%%%%%%%%%%%%%%%%%%%%%%%%%%%%%%%%%%%%%%%%%%%%%%%%%%%%%%%%%%%%%%%%

%%%%%%%%%%%%%%%%%%%%%%%%%%%%%%%%%%%%%%%%%%%%%%%%%%%%%%%%%%%%%%%%%%%%%%%%%%%%%%%%



% \section*{ACKNOWLEDGMENT}

% The preferred spelling of the word ÒacknowledgmentÓ in America is without an ÒeÓ after the ÒgÓ. Avoid the stilted expression, ÒOne of us (R. B. G.) thanks . . .Ó  Instead, try ÒR. B. G. thanksÓ. Put sponsor acknowledgments in the unnumbered footnote on the first page.

%%%%%%%%%%%%%%%%%%%%%%%%%%%%%%%%%%%%%%%%%%%%%%%%%%%%%%%%%%%%%%%%%%%%%%%%%%%%%%%%
% References are pulled from references.bib (ported from the CoRL 2026 submission).
\bibliographystyle{IEEEtran}
\bibliography{references}

\end{document}